%% file: main.tex
\documentclass{article}

\PassOptionsToPackage{numbers, compress}{natbib}

\usepackage{iclr2026_conference,times}

\iclrfinalcopy

\usepackage[utf8]{inputenc} %
\usepackage[T1]{fontenc}    %
\usepackage{hyperref}       %
\usepackage{url}            %
\usepackage{booktabs}       %
\usepackage{amsfonts}       %
\usepackage{nicefrac}       %
\usepackage{microtype}      %
\usepackage{xcolor}         %
\usepackage{xspace}
\usepackage{mathtools}
\usepackage{pifont}

\usepackage{amsmath}
\usepackage{amsthm}
\usepackage{algorithm}
\usepackage{algpseudocode}
\usepackage{graphicx}
\usepackage{subcaption}
\usepackage{tikz}
\usepackage{dsfont}
\usepackage{multirow}
\usepackage{wrapfig}
\usepackage[]{siunitx}

\usepackage{bm}
\usepackage{enumitem}

\usetikzlibrary{patterns}
\newtheorem{theorem}{Proposition}
\newtheorem{definition}{Definition}

\newcommand{\blackcircled}[1]{\tikz[baseline=(char.base)]{
    \node[shape=circle,fill=black,text=white,inner sep=1pt] (char) {\sffamily\small #1};}}

\newcommand{\codename}[0]{\textsc{DPQuant}\xspace}

\title{\codename: Efficient and Differentially-Private Model Training via Dynamic Quantization Scheduling}

\author{%
  \makebox[0.25\textwidth][l]{Yubo Gao} \\
  University of Toronto\\
  Toronto, Ontario\\
  \texttt{ybgao@cs.toronto.edu} \\
  \And
  \makebox[0.45\textwidth][l]{Renbo Tu} \\
  University of Toronto \\
  Toronto, Ontario \\
  \texttt{renbo.tu@mail.utoronto.ca} \\
  \AND
  \makebox[0.45\textwidth][l]{Gennady Pekhimenko} \\
  University of Toronto \\
  Toronto, Ontario \\
  \texttt{pekhimenko@cs.toronto.edu} \\
  \And
  \makebox[0.41\textwidth][l]{Nandita Vijaykumar} \\
  University of Toronto \\
  Toronto, Ontario \\
  \texttt{nandita@cs.toronto.edu} \\
}

\begin{document}

\maketitle

\vspace{-10pt}
\begin{abstract}
Differentially-Private SGD (DP-SGD) and its adaptive variant DP-Adam are powerful techniques to protect user privacy when using sensitive data to train neural networks. During training, converting model weights and activations into low-precision formats, i.e., \emph{quantization}, can drastically reduce training times, energy consumption, and cost, and is thus a widely used technique. In this work, we demonstrate for the first time that quantization causes significantly higher accuracy degradation in DP training compared to regular SGD. We observe that this is caused by noise injection, which amplifies quantization variance, leading to disproportionately large accuracy degradation.
To address this challenge, we present \codename, a dynamic quantization framework that adaptively selects a changing subset of layers to quantize at each epoch. Our method combines two key ideas that effectively reduce quantization variance: (i) \emph{probabilistic sampling} that rotates which layers are quantized every epoch, and (ii) \emph{loss-aware layer prioritization}, which uses a differentially private loss sensitivity estimator to identify layers that can be quantized with minimal impact on model quality. This estimator consumes a negligible fraction of the overall privacy budget, preserving DP guarantees. Empirical evaluations on ResNet18, ResNet50, and DenseNet121 across a range of datasets demonstrate that \codename consistently outperforms static quantization baselines, achieving near Pareto-optimal accuracy-compute trade-offs and up to \(2.21\times\) theoretical throughput improvements on low‑precision hardware, with less than $2\%$ drop in validation accuracy. We further show that our framework extends to DP-Adam with similar gains.
\end{abstract}

\section{Introduction}

Differentially Private Stochastic Gradient Descent (DP-SGD)~\citep{Abadi_2016} enables training neural networks on sensitive data while providing formal privacy guarantees. To improve the efficiency of such training on modern hardware, the use of  \textit{low-precision} arithmetic and data formats, i.e., \emph{quantization}, has gained widespread interest~\citep{gholami2021surveyquantizationmethodsefficient,jacob2017quantizationtrainingneuralnetworks}. Quantization can significantly  reduce the amount of computation and memory required, thus reducing the latencies, cost, and energy consumption during training and inference, often with little to no loss in model accuracy~\citep{micikevicius2022fp8formatsdeeplearning}. These benefits are especially important in resource-constrained settings such as federated learning with edge devices, where support for full-precision arithmetic is limited and compute budgets are constrained.

Modern accelerators, ranging from datacenter GPUs to mobile NPUs, are rapidly adopting \textit{ultra-low precision formats} such as FP8, INT4, or FP4. NVIDIA Blackwell~\citep{nvidia_blackwell_2024} provides 4$\times$ throughput for FP4 matmuls compared to FP16; AMD Instinct GPUs, Google TPUs and AWS Trainium support FP8~\citep{amd_instinct_mi300x_datasheet_2023,amd_rocm_precision_support_2025,vahdat2025ironwood,aws2025trainium}, and Qualcomm Hexagon supports INT4/INT8~\citep{qualcomm_ai_cores_2024}. Leveraging these compute capabilities in model training would enable significant performance and scalability improvements.

In this work, we observe that applying low-precision quantization directly to DP training often leads to \textit{significant accuracy degradation}, as severe as a $40\%$ drop. While non-DP training is typically robust to quantization, the gradient clipping and noise addition steps in DP-SGD/DP-Adam interact poorly with low-precision arithmetic leading to poor convergence as explained in Section~\ref{sec:degradation}.

Our goal is to develop an automatic mechanism to effectively quantize DP training, while minimizing the impact on model accuracy and the privacy budget. We observe that \textit{quantizing only a subset of layers} and \textit{selectively varying this subset every epoch} can preserve most of the efficiency gains from quantization while maintaining model accuracy. %
\codename uses two core techniques that are implemented in a \emph{differentially private} framework for dynamic quantization scheduling: 
\begin{enumerate}[leftmargin=2.5em]
    \item \textbf{Probabilistic layer sampling}, which rotates which layers are quantized every epoch to distribute quantization variance across the network, decreasing overall quantization variance;
    \item \textbf{Loss-aware prioritization}, which uses a loss sensitivity estimator to selectively quantize layers that have minimal impact on model accuracy.
\end{enumerate}

Specifically, in this work we make the following contributions:
\begin{enumerate}[leftmargin=2.5em]
    \item We are the first to demonstrate and explain the significant accuracy degradation that arises when employing ultra-low-precision (FP4/INT4) quantization with DP training (DP-SGD and DP-Adam), compared to the mild impact observed in non-private SGD and in 16-bit DP training.
    \item We introduce \codename, a differentially private lightweight mechanism that minimizes quantization-induced loss by (a) probabilistically sampling which layers to quantize every epoch and (b) prioritizing layers with lower sensitivity—while incurring only a negligible cost to the overall privacy budget.
    \item We demonstrate that \codename achieves near Pareto-optimal accuracy-speed tradeoffs across a range of compute and privacy budgets, outperforming static (fixed-layer) quantization baselines. 
\end{enumerate}

\section{Related Works}

\textbf{Post-training quantization (PTQ)}~\citep{ banner2019posttraining4bitquantizationconvolution, jacob2017quantizationtrainingneuralnetworks, nagel2021white} aims to accelerate inference through low-precision computations. A neural network is first trained in full-precision and then its weights are quantized. The conversion to quantized formats typically involves a small calibration dataset \citep{hubara2021accurate, nagel2019data} to allocate quantization bit-widths in different parts of the model and to perform bias correction. PTQ methods are orthogonal to this work since they do not optimize training.

\textbf{Quantization-aware training (QAT)}~\citep{krishnamoorthi2018quantizing} ameliorates the aforementioned accuracy loss by training in lower precision. However, QAT requires additional overhead such as hardware simulation \citep{wang2019haqhardwareawareautomatedquantization}, sensitivity estimation \citep{dong2019hawqhessianawarequantization}, and quantizer tuning \citep{esser2020learnedstepsizequantization, sakr2022optimal}, which typically cancels out raw bit-width speedups, making QAT suited for accelerating inference but not training \citep{chen2024efficientqat}.

\textbf{Gradient compression} \citep{lin2020deepgradientcompressionreducing,alistarh2017qsgdcommunicationefficientsgdgradient,alimohammadi2023lgrecolayerwiseadaptivegradientcompression,wen2017terngradternarygradientsreduce,shi2020scalabledistributedtrainingdeep} reduces communication costs by compressing gradients in distributed settings, either through sparsification \citep{stich2018sparsifiedsgdmemory, yu2017compressing} or low-rank approximation \citep{vogels2019powersgd, idelbayev2020low}. Notably, \citet{youn2023randomizedquantizationneeddifferential} combine quantization and the noising mechanism to achieve differential privacy while reducing communication. However, compression does not lower the arithmetic cost of training. In addition, these methods often rely on assumptions such as full gradient availability or error feedback accumulation \citep{karimireddy2019error}, which are difficult to satisfy under DP constraints. 

\textbf{Mixed-precision training}~\citep{choi2018pactparameterizedclippingactivation,zhou2018dorefanettraininglowbitwidth,ultralow,chmiel2024accurateneuraltraining4bit,micikevicius2022fp8formatsdeeplearning} aims to reduce training cost by operating on lower-precision data types (e.g., FP16, BF16, FP8, and FP4). Prior work has shown that 16-bit mixed-precision training (FP16/BF16) is compatible with DP-SGD with little accuracy loss~\citep{li2022largelanguagemodelsstrong,bu2023zeroredundancydistributedlearning}. However, pushing to ultra-low precision (FP4/INT4) under DP remains challenging: we show in Section~\ref{sec:degradation} that DP noise amplifies quantization variance at these bit-widths, causing significant accuracy degradation that does not arise at 16-bit precision.

\textbf{Quantization as a privacy mechanism.} Several works use quantization to provide or strengthen DP guarantees in federated or decentralized settings~\citep{kang2024effect,kim2022effects,wang2022quantizationenabledprivacyprotection,xiong2016randomized}. In these approaches, quantization is applied \emph{after} gradients are computed in full precision, so there is no effect on subsequent forward/backward passes. \codename instead quantizes activations and weights \emph{during} training to reduce arithmetic cost.

\textbf{Improved DP mechanisms.} A complementary line of work improves the DP algorithm itself for better utility at a given $\varepsilon$, whether through modified optimizers~\citep{wang2024dpadapterimprovingdifferentiallyprivate,Wei_2022,park2023differentiallyprivatesharpnessawaretraining,denisov2023improveddifferentialprivacysgd} or by reducing the trainable parameter space via pruning or layer selection~\citep{adamczewski2023differentialprivacymeetsneural}. These methods are orthogonal to \codename, which keeps the full dense model and instead optimizes how much computation can run in low precision.

\textbf{Gradient clipping optimizations for DP-SGD.} \citep{li2022largelanguagemodelsstrong,bu2022scalableefficienttraininglarge,lee2020scalingdifferentiallyprivatedeep,subramani2021enablingfastdifferentiallyprivate} optimize the per-sample gradient clipping for DP-SGD by eliminating redundant computation and increasing vectorized computation, leading to better hardware utilization. These works are orthogonal to \codename.

\section{Preliminaries}
\subsection{Differentially-Private DNN Training}
\label{sec:prelim-dpsgd}
We first recall the standard definition of differential privacy:

\begin{definition}[Differential Privacy, \citep{dwork2014algorithmic}]
A randomized algorithm $\mathcal{A}$ satisfies $(\varepsilon, \delta)$-differential privacy if for all adjacent datasets $D, D'$ differing on at most one example, and for all measurable sets $S$ in the output space, 
\[
\Pr[\mathcal{A}(D) \in S] \leq e^{\varepsilon} \Pr[\mathcal{A}(D') \in S] + \delta.
\]
\end{definition}

\begin{definition}[DP-SGD, \citep{Abadi_2016}]
Differentially Private Stochastic Gradient Descent (DP-SGD) is a variant of SGD that satisfies $(\varepsilon, \delta)$-differential privacy by clipping and perturbing per-example gradients. At each iteration $t$, the update rule is:
\[
\theta_{t+1} \leftarrow \theta_t - \eta \left( \frac{1}{|B|} \sum_{i \in B} \mathrm{clip}_C(\nabla \mathcal{L}(\theta_t, x_i)) + \mathcal{N}(0, \sigma^2 C^2 \mathds{1}) \right),
\]
where $B$ is a minibatch of training examples, $\mathrm{clip}_C(\cdot)$ scales the gradient to have $\ell_2$ norm at most $C$, and $\mathcal{N}(0, \sigma^2 C^2 \mathds{1})$ is Gaussian noise added to ensure privacy.
\end{definition}

\subsection{Quantization and Mixed Precision Training}

Modern hardware accelerators such as NVIDIA GPUs, Google TPUs, and Qualcomm Hexagon NPUs provide dedicated support for low-precision arithmetic, including fp16, bfloat16, and increasingly lower bitwidth formats like fp8, fp6, and fp4. These formats enable faster computation by reducing arithmetic complexity, memory footprint, and data transfer costs.

While prior work\citep{chmiel2024accurateneuraltraining4bit} has demonstrated that full training in low-bit formats (e.g., \texttt{fp4}) can often retain accuracy under standard SGD, extending these techniques to differentially private training remains challenging. The clipping and noise injection steps in DP-SGD amplify quantization errors and increase gradient variance, making DP training more sensitive to precision loss. Fully quantized DP-SGD thus often results in severe degradation unless carefully tuned.

\section{Degradation of DP Training from Quantization}
\label{sec:degradation}

\begin{figure}[h]
    \vspace{-10pt}
    \centering
    \begin{subfigure}{0.32\textwidth}
        \centering
        \includegraphics[width=\textwidth]{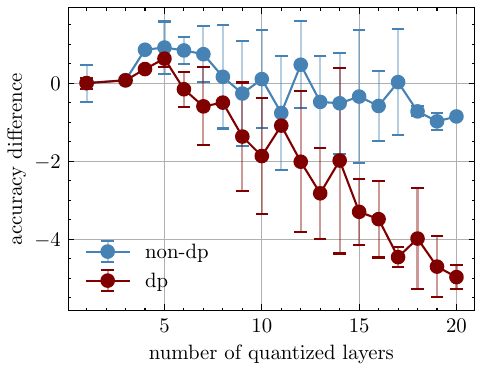}
        \caption{Accuracy loss due to quantization between DP and non-DP SGD\\}
        \label{fig:nondp_accuracy}
    \end{subfigure}\hfill
    \begin{subfigure}{0.32\textwidth}
        \centering
        \includegraphics[width=\textwidth]{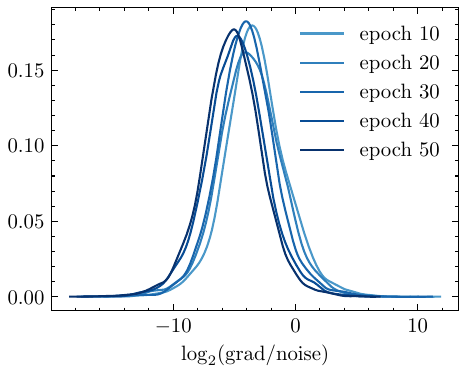}
        \caption{Distribution of gradient/noise ratios in \texttt{conv1} of ResNet18 -- Gradient mostly dominated by noise}
        \label{fig:noise_ratios}
    \end{subfigure}\hfill
    \begin{subfigure}{0.32\textwidth}
        \centering
        \includegraphics[width=\textwidth]{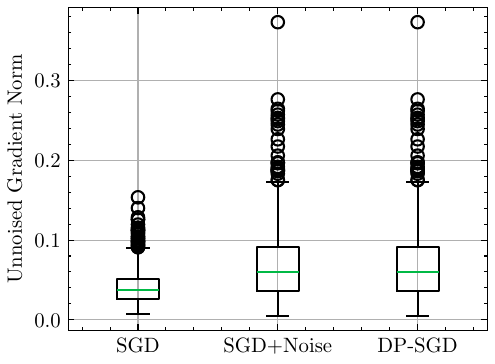}
        \vspace{-5pt}
        \caption{Distribution of gradient norms between SGD, SGD and only noise injection, and full DP-SGD}
        \label{fig:grad_norms}
    \end{subfigure}\hfill
    \caption{Comparing quantized SGD vs DP-SGD ResNet18 training on the German Traffic Sign Recognition Benchmark (GTSRB)~\citep{gtsrb} dataset}
    \label{fig:dp-vs-nondp}
\end{figure}

Figure~\ref{fig:dp-vs-nondp} presents a case study of ResNet18 trained on GTSRB, where the forward and backward convolution operators are quantized to FP4 to evaluate the effects on training. Figure~\ref{fig:nondp_accuracy} shows the accuracy loss compared to the non-quantized baseline for different degrees of quantization (in terms of the number of layers quantized), and the error bars represent the results when different subsets of layers are chosen for quantization, both for DP and non-DP training. 
For the non-DP SGD baseline, fully-quantized training results in only a modest accuracy drop of around 1\%. In contrast, DP-SGD experiences a much greater degradation, up to 5\%. Furthermore, the variance in performance due to different layers being quantized is substantially higher under DP-SGD. We observe similar trends in other neural networks and datasets (included in Appendix~\ref{ap:dp-degradation-other}). 

We hypothesize that the increased sensitivity to quantization can be attributed to noise injection in DP-SGD as follows. In iteration \(t\) in the DP-SGD training, the gradients $\mathbf{g}_t$ are first clipped to obtain $\bar{\mathbf{g}}_t$ where $||\bar{\mathbf{g}}_t||_2\leq C$ (section~\ref{sec:prelim-dpsgd}). Next, noise $\mathbf{n}_t\sim \mathcal{N}(0,\sigma^2C^2\mathds{1})$ is sampled, and finally the weights are updated using the sum of the noise and clipped gradient:
\begin{equation}
    \mathbf{w}_{t+1}=\mathbf{w}_t-\eta\,(\bar{\mathbf{g}}_t+\mathbf{n}_t)
    \label{eqn:weight-update}
\end{equation}
We assume the noise scale is $\sigma\in (0.5,10)$~\citep{Abadi_2016,de2022unlockinghighaccuracydifferentiallyprivate}, in line with common configurations reported in the DP-SGD literature. Since the standard deviation of the injected noise $\mathbf{n}_t$ is equal to the $2$-norm of the clipped gradients, the $\infty$-norm of the noise $\mathbf{n}_t$ (i.e. its largest component) is roughly on the same order as $||\mathbf{g}||_2$. Since in higher dimensions $||\bar{\mathbf{g}}||_2\gg||\bar{\mathbf{g}}||_\infty$ due to the $2$-norm growing much faster than the $\infty$-norm, combining we have:
\begin{equation}
    ||\mathbf n_t||_\infty 
    \;\approx\;||\bar{\mathbf g}_t||_2 
    \;\gg\;||\bar{\mathbf g}_t||_\infty
    \label{eqn:norm-dominate}
\end{equation}
This relation is also demonstrated empirically in Figure~\ref{fig:noise_ratios}, where on average the magnitude of the clipped gradient elements of $\bar{\mathbf{g}}$ is $2^5$ times smaller than that of the injected noise $\mathbf{n}$. 

The weight update (Equation~\ref{eqn:weight-update}) with noisy gradient updates amplify the norms of raw gradients in subsequent iterations. To show this, we first write the weight update as:
\begin{equation}
\Delta\mathbf w_t = 
\mathbf{w}_{t+1}-\mathbf{w}_t
=-\eta\,(\bar{\mathbf{g}}_t + \mathbf{n}_t)
\label{eq:weight-update}
\end{equation}
The weight update $\Delta \mathbf{w}_t\approx \eta\, \mathbf{n}_t$ due to \(\mathbf n_t\) being much larger in norm than $\bar{\mathbf{g}}_t$, and thus:
\begin{equation}
||\Delta\mathbf w_t||_\infty
\approx \eta\,||\mathbf n_t||_\infty=\mathcal{O}(||\bar{\mathbf{g}}_t||_2)
\label{eq:weight-update-asymptotic}
\end{equation}
where the asymptotic equality holds due to Equation~\ref{eqn:norm-dominate}. Assuming $L$-Lipschitz‐smoothness of the loss with respect to the gradients:
\begin{equation}
||\mathbf g_{t+1}-\mathbf g_t||_\infty
\;\le\;
L\,||\mathbf{w}_{t+1}-\mathbf{w}_t||_\infty =
L\,||\Delta\mathbf w_t||_\infty
\label{eq:lipschitz}
\end{equation}
We now show that the $\infty$-norm of the raw gradients (i.e. before clipping and noising) of the next iteration is bounded by $||\bar{\mathbf{g}}_t||_2$ using the inverse triangle inequality with Equation~\ref{eq:weight-update-asymptotic} and \ref{eq:lipschitz}:
\begin{equation}
||\mathbf{g}_{t+1}||_\infty \geq ||\mathbf{g}_{t+1}-\mathbf{g}_t||_\infty - ||\mathbf{g}_t||_\infty =\mathcal{O}(||\mathbf{g}_t||_2)
\label{eq:infty-bound-next-grad}
\end{equation}
Equation~\ref{eq:infty-bound-next-grad} shows that elements of the raw gradients in the next iteration are bounded by the much larger $\mathcal{O}(||\bar{\mathbf{g}}_t||_2)$ rather than the usual $\mathcal{O}(||\bar{\mathbf{g}}||_\infty)$ in normal SGD. As a result, we expect elements of the raw gradients in DP-SGD to be larger in magnitude than in non-DP training. 

Notably, the batch size has a negligible effect on norms, even though a larger batch size leads to a smaller variance of the stochastic gradients. Therefore, we omit it in this analysis; we include further discussions and empirical evaluations of this in appendix~\ref{ap:batch-size}.

To show this empirically, we plot the norms of intermediate gradients under both SGD and DP-SGD using the same hyperparameters in Figure \ref{fig:grad_norms}, where the DP-SGD intermediate raw gradients are $2\times$ larger in the average and worst case. This phenomenon has also been observed in prior work~\citep{du2022dynamicdifferentialprivacypreservingsgd}, showing an even larger gap than what we observe in gradient norms later in training.

We now demonstrate that the much larger raw gradients under DP-SGD result in much higher quantization variance.
\begin{theorem}
    Let $q:\mathbb{R}^n\to\mathbb{R}^n$ be an unbiased (i.e. $\mathbb{E}[q(\mathbf{x})]=\mathbf{x}$) and scale invariant (i.e. $q(\lambda \mathbf{x})=\lambda q(\mathbf{x})$) quantizer. Assume $q(\mathbf{x})$ quantizes values onto some finite grid. Let $\mathbf{x}$ be sampled from an absolutely continuous distribution. Then the quantizer variance $\operatorname{Var}(q(\mathbf{x}))=\Theta(||\mathbf{x}||_\infty^2)$. Proof: See Appendix ~\ref{ap:proof-prop-1}.
    \label{prop:quantizer_noise}
\end{theorem}
Using Prop. \ref{prop:quantizer_noise}, we can more precisely express the variance of the quantization as follows:
\begin{align*}
\text{(under DP-SGD)} &\qquad \mathrm{Var}\bigl(q(\mathbf g_{t+1})\,|\,\mathbf{g}_{t+1}\bigr) =\mathcal{O}\bigl(||\mathbf g_{t+1}||_\infty^2\bigr) =\mathcal{O}\bigl(||\mathbf g_t||_2^2\bigr) \\
\text{(under SGD)} &\qquad \mathrm{Var}\bigl(q(\mathbf g_{t+1})\,|\,\mathbf{g}_{t+1}\bigr) =\mathcal{O}\bigl(||\mathbf g_t||_\infty^2\bigr)
\end{align*}
The quantization variance above is \emph{in addition} to the existing variance of the stochastic gradients, as well as noise injected by DP-SGD. In higher dimensions, \(||\mathbf g_t||_2\gg||\mathbf g_t||_\infty\), quantization contributes much more variance to the gradients, hence leads to slower and less reliable convergence~\citep{NIPS2013_ac1dd209} and accuracy degradation. To address this challenge, we aim to reduce the quantization-induced variance. 

\paragraph{Extension to DP-Adam/DP-AdamW.} The above argument for DP-SGD relies on DP noise dominating the raw gradients, which leads to disproportionate quantization variance. In DP-Adam, noise is added to the clipped gradients first, and Adam's preconditioner then rescales the noisy gradients coordinate-wise. Since the preconditioner applies the same scaling to both signal and noise, it preserves the relative scale between them, so quantized DP-Adam incurs the same quantization overhead as DP-SGD.

We can reason about this more directly by considering the per-coordinate signal-to-noise ratio. In DP-Adam, each noisy gradient coordinate is $\tilde{g}_i = g_i + n_i$, where $g_i$ is the true gradient and $n_i\sim\mathcal{N}(0,\sigma^2C^2)$. The preconditioner divides by a running second-moment estimate, so at stationarity the preconditioned coordinate is $u_i\approx (g_i+n_i)/\sqrt{g_i^2+\sigma^2C^2}$ with $\operatorname{Var}(u_i)\leq 1$. 

In the strong DP regime (where $\sigma C\gg ||\mathbf{g}||_2 \gg |g_i|$), the mean (i.e. signal) is:
\begin{equation}
    |\mathbb{E}[u_i]|=\frac{|g_i|}{\sqrt{g_i^2+\sigma^2C^2}}\approx \frac{|g_i|}{\sigma C} \leq \frac{1}{\sigma}
\end{equation}
where the last inequality is due to $|g_i| \leq ||\mathbf{g}||_2 \leq C$.
We define the per-coordinate SNR in the preconditioned space, which is directly comparable to non-preconditioned DP-SGD:
\begin{equation}
    \text{SNR}_i \coloneqq \frac{(\mathbb{E}[u_i])^2}{\operatorname{Var}(u_i)} \approx \frac{g_i^2/(\sigma^2 C^2)}{1} = \frac{g_i^2}{\sigma^2 C^2}.
\end{equation}
Thus the SNR in DP-Adam matches that of DP-SGD. Therefore, DP turns each preconditioned coordinate into unit-variance noise with very small mean. A fixed-precision quantizer on $u_i$ then adds zero-mean noise whose variance does not depend on this small mean, so the signal-to-quantization-noise ratio behaves like $g_i^2/(\sigma^2 C^2)$ and deteriorates as privacy noise increases.

\section{\codename: Our Proposed Solution}

\subsection{Part I: Probabilistic Layer Sampling}
\label{sec:part1}
Each time we perform randomized and unbiased quantization on a layer, we introduce additional variance to its gradient updates. While this added variance might be acceptable in standard training, it results in a significant performance degradation under DP-SGD, where gradient stability is already challenged by injected noise.

Suppose a layer is quantized with probability $p$, we let $\mathbf{g}_\text{fp}$ denote its full precision gradients and $\mathbf{g}_\text{quant}$ to be its gradients computed under quantization. By Section~\ref{sec:degradation}, quantization incurs additional variance, hence $\operatorname{Var}(\mathbf{g}_\text{fp})\leq \operatorname{Var}(\mathbf{g}_\text{quant})$. The \emph{expected} gradient variance is:
\[
    \mathbb{E}\left(\operatorname{Var}( \mathbf{g} )\right) = (1 - p)\operatorname{Var}(\mathbf{g}_\text{fp}) + p\operatorname{Var}(\mathbf{g}_\text{quant}) \leq \operatorname{Var}(\mathbf{g}_\text{quant})
\]

\begin{wrapfigure}{r}{0.4\textwidth}
  \centering
  \begin{minipage}{\linewidth}
    \vspace{-20pt}
    \includegraphics[width=\textwidth]{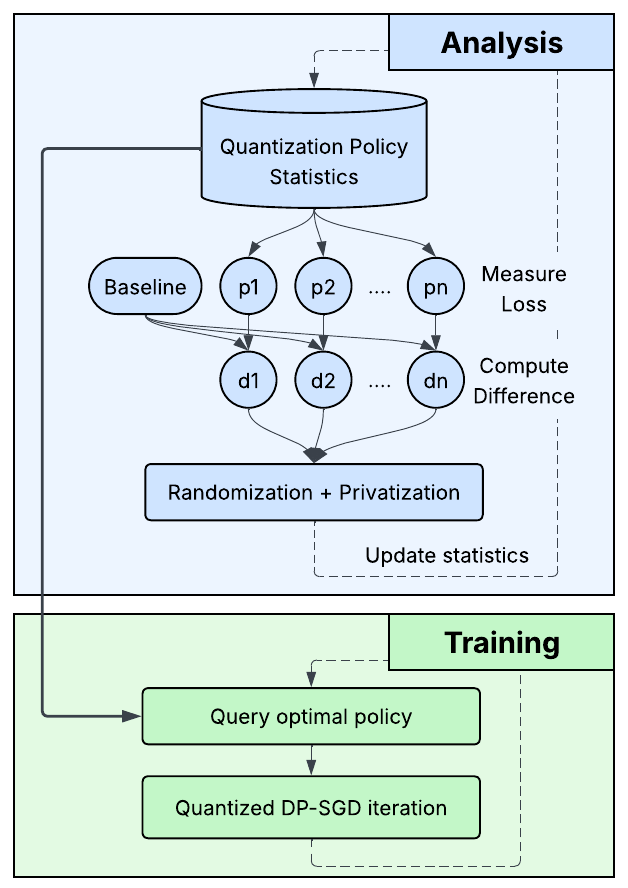}
    \caption{\codename system overview}
    \label{fig:sys_overview}
    \vspace{-20pt}
  \end{minipage}
\end{wrapfigure}

From this it follows that whenever $p<1$ -- that is, when only a subset of layers is quantized at each epoch -- the average quantization‐induced variance is strictly lower than in full quantization. Furthermore, by rotating which layers are quantized every epoch, no single layer repeatedly incurs the full quantization variance, and hence their expected variance remains smaller than $\operatorname{Var}(\mathbf{g}_\text{quant})$.

\subsection{Part II: Loss-Aware Layer Prioritization}
\label{sec:part2}

Not all layers contribute equally to model performance. Intuitively, we prefer to retain higher precision in layers that have a greater impact on the loss or accuracy. Given a constrained quantization budget, our goal is to prioritize quantization in lower-impact layers, thereby minimizing the overall loss in model quality.

We define a quantization policy $p$ to be the set of layers to compute under quantization. We define \(R(p)\) as the expected \emph{loss increase} incurred by applying quantization policy $p$ instead of full precision:
\[
R(p) \;:=\; \mathbb{E}_{D}\bigl[\mathcal{L}\bigl(M_{p}(D)\bigr)\;-\;\mathcal{L}\bigl(M_{\mathrm{fp32}}(D)\bigr)\bigr].
\]
Our goal is to find policies that minimally increase loss (or equivalently, a policy $p$ with small $R(p)$). We note that a key challenge in evaluating $R(p)$ for a given policy $p$ is that the expectation is taken over the full private dataset $D$. This makes direct computation both expensive and incompatible with tight privacy guarantees. To address this, we instead estimate $R(p)$ by subsampling $D$ and running a limited number of DP-SGD iterations under policy $p$ to obtain a proxy loss, then the same training iterations are done to obtain the baseline full-precision loss, and the difference is used as an empirical estimate of the loss impact.

Since this quantity is computed on the private training dataset $D$, any estimation of $R(\cdot)$ must be performed in a differentially private manner, and therefore consumes part of the overall privacy budget. Any computation using the private data incurs a privacy cost that must be accounted for to ensure that the privacy budget is not exceeded. We outline how \codename privatizes and accounts for loss measurement in Section \ref{sec:priv}.

\subsection{Dynamic Layer Selection for Quantized DP Training}
Building on the insights from the previous sections, we design a dynamic layer selection strategy for quantized DP training that combines: (i) probabilistic sampling of quantized layers to reduce variance (Section~\ref{sec:part1}), and (ii) loss-aware prioritization to preserve performance by avoiding quantization of high-impact layers (Section~\ref{sec:part2}).

Let $R(l_i)$ denote the estimated quantization loss impact of layer $i$. We define the probability of selecting layer $i$ for quantization as:
\[
    \pi_i \coloneqq \frac{\exp(-\beta R(l_i))}{\sum_{j=1}^n \exp(-\beta R(l_j))}, \qquad \text{for } i = 1, \ldots, n,
\]
where $\beta > 0$ is a scaling parameter that controls how strongly we prioritize low-impact layers.

As outlined in Figure~\ref{fig:sys_overview}, to quantize $k$ out of $n$ layers at each epoch, we sample a subset \emph{without replacement} according to the distribution $\{\pi_i\}$. This allows us to adaptively choose the least sensitive layers for quantization, while still randomly rotating layers with similar loss impact to minimize variance over time. This selection procedure is detailed in Appendix~\ref{ap:select-targets}. \codename provides a set of tunable parameters that govern the frequency of the analysis, as well as other privacy parameters. 

\subsection{Privacy Accounting}
\label{sec:priv}

Our method begins by measuring loss differences on each user’s private dataset. Specifically, we compute $\mathcal{L}(M(D))$ which requires inspecting raw data and inherently risks exposing sensitive information if released directly. Without these privacy-preserving measures, simply publishing the loss-difference measurements compromises the privacy guarantee DP-SGD provides. 

\begin{definition}[Sampled Gaussian Mechanism (SGM), \citep{mironov2019renyidifferentialprivacysampled}] 
Let $f$ be a function that maps subsets of a dataset $S$ to $\mathbb{R}^d$. The Sampled Gaussian Mechanism, denoted $\mathrm{SG}_{q, \sigma}$, is defined with sampling rate $0 < q \leq 1$ and noise parameter $\sigma > 0$ as:
\[
\mathrm{SG}_{q, \sigma}(S) \coloneqq f\left(\{x \in S : x \text{ is independently sampled with probability } q\}\right) + \mathcal{N}(0, \sigma^2 \mathbb{I}^d),
\]
where each element in $S$ is independently included with probability $q$, and $\mathcal{N}(0, \sigma^2 \mathbb{I}^d)$ denotes $d$-dimensional isotropic Gaussian noise with variance $\sigma^2$ per coordinate.
\end{definition}

To protect privacy, we frame this loss computation as a Sampled Gaussian Mechanism (SGM): we draw a random subsample of \(D\), clip the resulting loss value to bound sensitivity, and then add Gaussian noise of scale \(\sigma\). These operations correspond to step 3 of Algorithm~\ref{alg:compute_regret_concise}.
\input{sections/alg1_small}

\begin{theorem}
    Algorithm \ref{alg:compute_regret_concise} is a Sampled Gaussian Mechanism (SGM) with sample rate $q=|B|/|D|$ and noise scale $\sigma=\sigma_\text{measure}$. Proof: See Appendix~\ref{ap:proof-prop-2}.
    \label{prop:sgm}
\end{theorem}

To account for the privacy cost we incur by performing the analysis in Algorithm~\ref{alg:compute_regret_concise}, we rely on Opacus’s privacy accountants~\citep{yousefpour2022opacususerfriendlydifferentialprivacy}. This is for two reasons: First, these accountants measure the cumulative privacy loss of SGMs~\citep{makni2025optimizationframeworkdifferentiallyprivate}, where by Prop.~\ref{prop:sgm} we can reuse its implementation. Second, by leveraging the advanced composition theorem~\citep{Abadi_2016}, we obtain a much tighter upper bound on the total privacy expenditure incurred by both the DP-SGD training process and any subsequent analyses performed under the same privacy budget. We explain this in more detail in Appendix~\ref{ap:opacus-accounting}.

\begin{figure}[h]
    \centering
    \begin{subfigure}{0.44\textwidth}
        \centering
        \includegraphics[width=\textwidth]{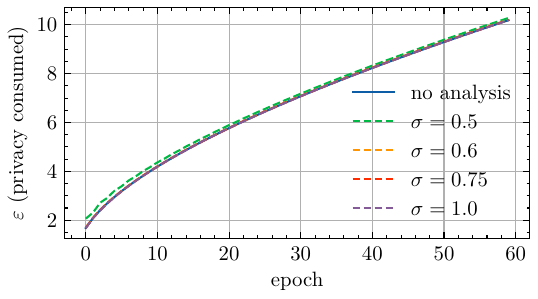}
        \caption{privacy consumption of analysis + training}
    \end{subfigure}\hfill
    \begin{subfigure}{0.44\textwidth}
        \centering
        \includegraphics[width=\textwidth]{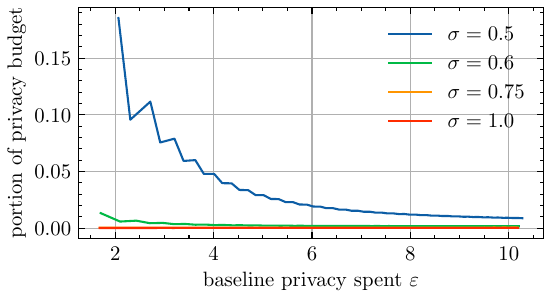}
        \caption{Fraction of privacy spent on analysis}
    \end{subfigure}\hfill
    \caption{Privacy cost of analysis for ResNet18/GTSRB; performing analysis every 2 epochs}
    \vspace{-10pt}
    \label{fig:privacy_consumption}
\end{figure}

In Figure~\ref{fig:privacy_consumption}, we report the cumulative privacy loss from both the analysis and training components across various configurations. Our results empirically demonstrate that the privacy cost of analysis is negligible compared to training, and does not meaningfully affect the quality of the resulting model.

\section{Evaluation}
\textbf{Models and Datasets.} We evaluate our approach on commonly used neural networks for differentially private training~\citep{jagielski2020auditingdifferentiallyprivatemachine,de2022unlockinghighaccuracydifferentiallyprivate}: ResNet18~\citep{he2015deepresiduallearningimage}, ResNet50 and DenseNet121~\citep{huang2018denselyconnectedconvolutionalnetworks} from the \texttt{torchvision}~\citep{torchvision2016} library. We also test BERT~\citep{devlin2019bertpretrainingdeepbidirectional}, trained with DP-AdamW. We additionally evaluate with DP-Adam in Appendix~\ref{ap:dp-adam}. These models are trained on the Extended MNIST~\citep{cohen2017emnistextensionmnisthandwritten}, German Traffic Sign Recognition Benchmark (GTSRB)~\citep{gtsrb}, CIFAR-10~\citep{krizhevsky2009learning} and SNLI~\citep{bowman2015largeannotatedcorpuslearning} datasets.

\textbf{Implementation.} \codename is implemented on top of Opacus~\citep{yousefpour2022opacususerfriendlydifferentialprivacy}, a DP training framework which provides Poisson sampling, gradient clipping, and noising. The \codename parameters can be found in Appendix~\ref{ap:dpquant-hyperparams}. %

\textbf{Low Precision Format.} For low precision computations, we used the LUQ-FP4~\citep{chmiel2024accurateneuraltraining4bit} format, the highest-performing 4-bit quantization format. LUQ-FP4 uses a 4-bit representation of floating point numbers, consisting of 1 sign and 3 exponent bits. In Appendix~\ref{ap:other-quantizers}, we evaluate \codename on other low-precision formats including FP8 and 4-bit uniform quantization.

\vspace{-5pt}

\subsection{Quantization-quality trade-off}
Quantizing more layers proportionally increases the speed of training. However, it also increases the accuracy degradation in DP-SGD training. Thus, there is a \emph{speed-accuracy} trade-off depending on the number of layers quantized. For a given number of quantized layers, the resulting model accuracy can significantly vary depending on which layers are quantized at any given epoch. \codename\ aims to automatically identify the subset of layers for each epoch that provides the best accuracy, assuming a certain number of layers are quantized. We refer to the desired number of quantized layers as ``computational budget'' because it determines the speed and compute resources needed. 

\begin{figure}[t]
    \centering
    \includegraphics[width=0.95\textwidth]{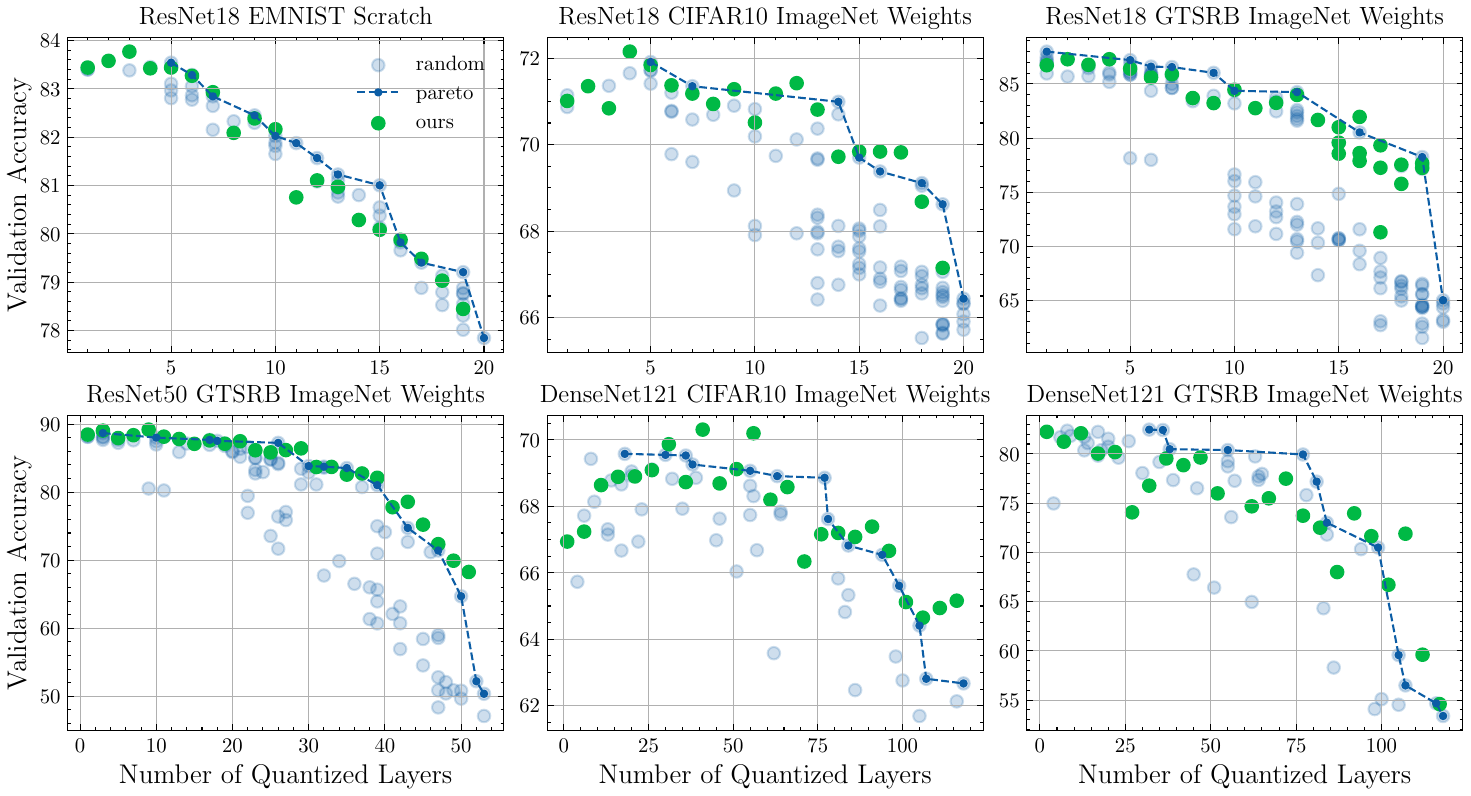}
    
    \caption{Comparing policies generated by \codename to the speed-accuracy Pareto front}
    \label{fig:pareto}
    \vspace{-10pt}
\end{figure}

In Figure \ref{fig:pareto}, we sample $\approx 50$ random subsets of layers to execute in fp4. We plot the empirical Pareto front using these sampled measurements, in addition to the resulting validation accuracy when using \codename's scheduling technique for a given computational budget. 

We make two observations. First, we note that randomly selecting the quantized layers can lead to significant loss in accuracy, as much as $40\%$. Second, \codename\ generates scheduling configurations that provide validation accuracy close to the Pareto-front for all evaluated networks and datasets.

\subsection{Sensitivity to Privacy Budget}
\input{sections/table_eps}

We compared our method to the baseline for two privacy budgets $\varepsilon=4$ and $\varepsilon=8$. In Table \ref{tbl:diff_eps} we report the validation accuracy for different privacy budgets. For ResNet18/50, we obtain these values by truncating the training at the respective privacy budgets (i.e. without additional hyperparameter tuning), and select baseline data points with larger $\varepsilon$ than ours wherever possible. 

In most cases, \codename outperforms the baseline performance by at least 1 standard deviation whilst not exceeding the privacy budget. In particular, despite the privacy cost of analysis being more dominant during the $\varepsilon=4$ case, \codename produces near-optimal quantization schedules, demonstrating its robustness with respect to $\varepsilon$. 

We have also evaluated \codename on extremely small privacy budgets (e.g. $\varepsilon=1$). In Appendix~\ref{ap:small-privacy-budget} we show that \codename still demonstrates the same benefits under this setting.

\subsection{Ablation Study}

\begin{figure}[h]
    \centering
    \begin{minipage}{0.38\textwidth}
        \centering
        \includegraphics[width=\linewidth]{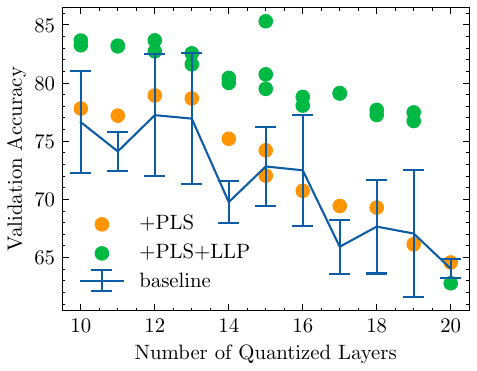}
        \captionof{figure}{Ablation study, PLS: probabilistic layer selection, LLP: loss-aware layer prioritization}
        \label{fig:ablation}
    \end{minipage}\hfill
    \begin{minipage}{0.58\textwidth}
        \centering
        \includegraphics[width=\linewidth]{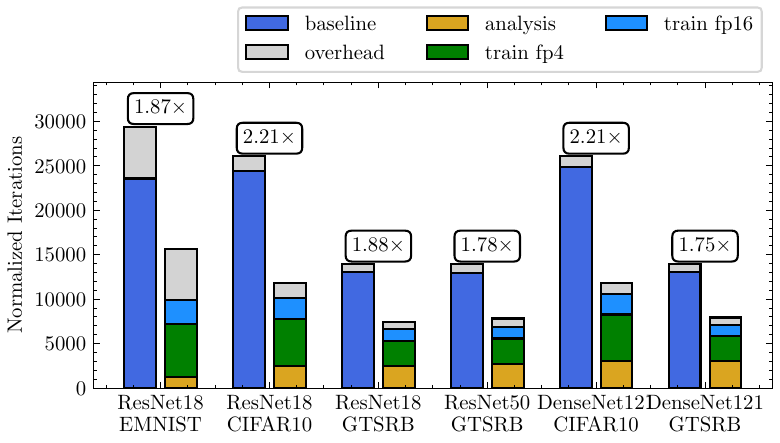}
        \captionof{figure}{Theoretical speedups for \codename{} assuming $90\%$ of the layers are quantized.}
        \label{fig:perf}
    \end{minipage}
    \vspace{-10pt}
\end{figure}

To better understand the contributions of the two approaches, we compared our approach (probabilistic layer sampling + loss-aware layer prioritization) with probabilistic layer sampling (PLS) alone. In Figure \ref{fig:ablation}, we observe that PLS consistently performs better than the baseline where the quantized layers are selected statically. However, there is still a large gap between PLS and the \emph{best-performing} layer selections, suggesting that some crucial layers are consistently being subjected to quantization which significantly degrades the quality of trained models.

When PLS is combined with loss-aware layer prioritization, the layers crucial to model training are left in full precision, even when most of the layers are quantized. The benefits of prioritization begins to surface as the proportion of quantized layers increase, as the critical layers have a larger probability of being quantized in the randomized baselines. Furthermore, we observe that the best training outcomes are achieved by combining both approaches. We include more details in Appendix~\ref{ap:beta-sensitivity}.

\subsection{Theoretical Speedup}
As hardware with support for FP4 MatMuls and Conv2D (e.g., NVIDIA Blackwell) are not yet widely available, we are unable to evaluate the speed benefits of quantization with \codename. Instead, we use estimates from prior work, along with performance statistics published by NVIDIA~\citep{nvidia_blackwell_2024} to estimate speedups. Based on profiling on existing hardware, we estimate that FP4 can provide a $4\times$ speedup over the FP16 baseline. Separately, prior works~\citep{ultralow,choi2018pactparameterizedclippingactivation,Abdolrashidi_2021} report a $4-7.3\times$ speedup when using FP4 on supported hardware. To remain conservative, we use the lower bound ($4\times$) in our estimates. We assume matrix multiplications, convolutions, and element-wise operations can be accelerated $4\times$, and characterize the total runtime as a linear compute cost model:
\[
    T_\text{ours}=T_\text{analysis}+(1-p+p/4)(T_\text{train baseline}-T_\text{overhead})+T_\text{overhead}
\]
where $T_\text{analysis}$ is the time taken by algorithm \ref{alg:compute_regret_concise}, and $T_\text{overhead}$ captures the time taken by operations that do not have performance benefits from low precision (details in appendix~\ref{ap:speedup}).

We show our speedups in Figure~\ref{fig:perf}. Quantized training with \codename is $1.75\times$ to $2.21\times$ faster than the fp16 baseline. In particular, the loss-aware prioritization mechanism in \codename incurs minimal runtime overhead, which is crucial to preserve the performance gains of fp4 computation.

\section{Conclusion}
In this paper, we introduce \codename, a mechanism for efficient quantized DP training. We observe that existing quantized training techniques can significantly degrade the accuracy of models trained with DP-SGD and DP-Adam, and provide a detailed analysis demonstrating the amplified quantization error. To address this challenge, \codename employs techniques to dynamically select layers to quantize such that the impact of quantization on model accuracy is minimized. \codename itself is a differentially private mechanism that incurs only small privacy cost. We empirically demonstrate that \codename achieves near-optimal compute-to-accuracy tradeoffs during quantized training, generalizes to different models, datasets and privacy budgets, and can provide up to $2.21\times$ speedup while minimally impacting accuracy. \codename enables efficient and practical differentially-private training for both centralized and distributed training deployments. 

\newpage
\bibliographystyle{plainnat}
\bibliography{ref}

\input{sections/appendix}

\end{document}

%% file: sections/alg1_small.tex
\begin{algorithm}[]
\caption{\textsc{ComputeLossImpact}}
\label{alg:compute_regret_concise}
\begin{algorithmic}[1]
\State \textbf{Input:} $P$ (policies), $B$ (batches), $R$ (iterations), $\alpha$ (decay), $\mathcal{C}$ (norm), $\sigma$ (noise)
\State Let $p_0$ be the baseline policy (no quantization)
\State Initialize a map for average losses, $\bar{\ell}$

\For{each $p \in P \cup \{p_0\}$} \Comment{\blackcircled{1} Compute avg. loss for baseline and all policies}
    \State $\text{total\_loss} \gets 0$
    \For{$i = 1$ to $R$}
        \State \textsc{RestoreModel}()
        \For{each $(x, y) \in B$}
            \State With policy $p$, run $\textsc{DPSGD-Update}(M, \text{loss}(M(x), y))$
        \EndFor
        \State $\text{total\_loss} \gets \text{total\_loss} + \frac{1}{|B|} \sum_{(x, y) \in B} \text{loss}(M(x), y)$
    \EndFor
    \State $\bar{\ell}[p] \gets \text{total\_loss} / R$
\EndFor

\State $R[p] \gets \bar{\ell}[p] - \bar{\ell}[p_0]$ for all $p \in P$ \Comment{\blackcircled{2} Compute loss differences from baseline}
\State $\mathbf{R} \gets [R[p_1], \ldots, R[p_k]]$
\State $\hat{\mathbf{R}} \gets \mathbf{R} \cdot \min\left(1, \frac{\mathcal{C}}{||\mathbf{R}||_2}\right) + \mathcal{N}(0, \sigma^2 \mathcal{C}^2\mathds{1})$ \Comment{\blackcircled{3} Privatize differences}

\State \textsc{UpdatePrivacy}(
rate$\,{=}\,|B|/|D|$, steps$\,{=}\,1$, noise\_scale$\,{=}\,\sigma$)

\For{each $p \in P$} 
    \State $L[p] \gets (1 - \alpha) \cdot L[p] + \alpha \cdot \hat{\mathbf{R}}[p]$ \Comment{\blackcircled{4} Update Policy EMA}
\EndFor
\State \Return $L$
\end{algorithmic}
\end{algorithm}

%% file: sections/table_eps.tex
\newcommand{\mustd}[2]{%
    \SI[round-mode=places,round-precision=2]{#1}{\relax} \,$\pm$\, \textcolor{gray}{\SI[round-mode=places,round-precision=2]{#2}{\relax}}%
}

\newcommand{\rn}[1]{\SI[round-mode=places,round-precision=2]{#1}{\relax}}
\newcommand{\brn}[1]{\textbf{\SI[round-mode=places,round-precision=2]{#1}{\relax}}}

\begin{table}[h]
\centering
\resizebox{\textwidth}{!}{%
\begin{tabular}{llcccccccccc}
\toprule
\multirow{2}{*}{Model} & \multirow{2}{*}{Dataset} & \multirow{2}{*}{\shortstack{Percent \\Quantized}} & \multicolumn{4}{c}{$\varepsilon = 4$} & \multicolumn{4}{c}{$\varepsilon = 8$} \\
\cmidrule(lr){4-7} \cmidrule(lr){8-11}
 & & & \text{Baseline} & \text{$\varepsilon$} & \text{Ours} & \text{$\varepsilon$} & \text{Baseline} & \text{$\varepsilon$} & \text{Ours} & \text{$\varepsilon$} \\
\midrule
\multirow{9}{*}{ResNet18} 
 & \multirow{3}{*}{EMNIST}   & 0.5  & \mustd{81.27234}{1.292118} & \rn{3.142105} & \brn{82.164894} & \rn{3.043234} & \multicolumn{4}{c}{--} \\
 &                           & 0.75 & \mustd{80.514628}{0.369859} & \rn{3.012279} & \rn{80.085106} & \rn{3.043234} & \multicolumn{4}{c}{--} \\
 &                           & 0.9  & \mustd{78.817376}{0.303953} & \rn{3.012279} & \brn{79.031915} & \rn{3.043234} & \multicolumn{4}{c}{--} \\
\cmidrule{2-11}
 & \multirow{3}{*}{GTSRB}    & 0.5  & \mustd{42.336698}{5.529033} & \rn{4.010614} & \brn{49.08947} & \rn{3.987328} & \mustd{69.060768}{5.625092} & \rn{8.010063} & \brn{76.745843} & \rn{7.994121} \\
 &                           & 0.75 & \mustd{39.976247}{3.992324} & \rn{4.010614} & \brn{42.618105} & \rn{3.961811} & \mustd{63.617049}{5.588763} & \rn{8.010063} & \brn{70.071259} & \rn{7.988856} \\
 &                           & 0.9  & \mustd{37.936885}{2.231238} & \rn{4.010614} & \brn{39.477435} & \rn{3.987328} & \mustd{57.488972}{4.455325} & \rn{8.010063} & \brn{67.66825} & \rn{7.994121} \\
\cmidrule{2-11}
 & \multirow{3}{*}{CIFAR-10} & 0.5  & \mustd{64.365}{1.415121} & \rn{4.056089} & \brn{65.39} & \rn{3.935441} & \mustd{69.26}{1.462942} & \rn{7.118166} & \brn{70.51} & \rn{7.167737} \\
 &                           & 0.75 & \mustd{62.17}{0.607124} & \rn{4.056089} & \brn{63.57} & \rn{3.935441} & \mustd{67.796667}{0.807543} & \rn{7.118166} & \brn{69.84} & \rn{7.167737} \\
 &                           & 0.9  & \mustd{61.0925}{1.661899} & \rn{4.056089} & \rn{61.22} & \rn{3.935441} & \mustd{67.21375}{1.242623} & \rn{7.118166} & \brn{68.68} & \rn{7.167737} \\
\midrule
 \multirow{3}{*}{ResNet50} 
 & \multirow{3}{*}{GTSRB}    & 0.5  & \mustd{38.758907}{8.157116} & \rn{4.010614} & \brn{42.106097} & \rn{3.987328} & \mustd{75.985748}{7.332935} & \rn{8.010063} & \brn{80.229612} & \rn{7.994121} \\
 &                           & 0.75 & \mustd{29.482713}{4.718487} & \rn{4.010614} & \brn{33.665875} & \rn{3.987328} & \mustd{58.131433}{8.504299} & \rn{8.010063} & \brn{69.034046} & \rn{7.994121} \\
 &                           & 0.9  & \mustd{24.779625}{2.668256} & \rn{4.010614} & \brn{29.002375} & \rn{3.987328} & \mustd{47.403009}{7.226648} & \rn{8.010063} & \brn{59.865400} & \rn{7.994121} \\
\midrule
\multirow{6}{*}{DenseNet121} 
 & \multirow{3}{*}{GTSRB\footnotemark{}}    & 0.5  & \mustd{54.096307}{5.576742} & \rn{4.056767} & \rn{55.380048} & \rn{3.973971}  & \mustd{65.472090}{5.421880} & \rn{8.010063} & \brn{71.053048} & \rn{7.928890} \\
 &                           & 0.75 & \mustd{44.601742}{5.056124} & \rn{4.056767} & \brn{47.359462} & \rn{3.973971} & \mustd{56.143112}{7.567057} & \rn{8.010063} & \brn{63.301663} & \rn{7.928890} \\
 &                           & 0.9  & \mustd{40.521686}{2.834334} & \rn{4.056767} & \brn{44.148852} & \rn{3.973971} & \mustd{51.060966}{5.407077} & \rn{8.010063} & \brn{52.604909} & \rn{7.928890} \\
\cmidrule{2-11}
 & \multirow{3}{*}{CIFAR-10\footnotemark[\value{footnote}]} & 0.5  & \mustd{59.221000}{1.154435} & \rn{4.025226} & \brn{61.080000} & \rn{3.968785} & \mustd{67.960909}{0.925489} & \rn{7.118166} & \brn{68.960000} & \rn{7.275078} \\
 &                           & 0.75 & \mustd{56.430000}{1.723459} & \rn{4.025226} & \brn{60.310000} & \rn{3.968785} & \mustd{64.808000}{1.705004} & \rn{7.118166} & \brn{66.480000} & \rn{7.275078} \\
 &                           & 0.9  & \mustd{55.175556}{1.382978} & \rn{4.025226} & \brn{58.890000} & \rn{3.968785} & \mustd{63.031111}{1.690846} & \rn{7.118166} & \brn{65.130000} & \rn{7.275078} \\
 \midrule
 \multirow{2}{*}{BERT} 
 & \multirow{2}{*}{SNLI}    & 0.5  &  & & & & \mustd{62.54}{4.54} & \rn{7.48} & \brn{67.80} & \rn{7.48} \\
 &                           & 0.75 &  &  &  &  & \mustd{52.04}{3.95} & \rn{7.48} & \brn{63.61} & \rn{7.48} \\
\bottomrule
\end{tabular}%
}%
\caption{Model quality across datasets and privacy levels.}
\label{tbl:diff_eps}
\vspace{-15pt}
\end{table}
\footnotetext{Batch size decreased to improve convergence under $\varepsilon=4$.}

%% file: sections/appendix.tex
\newpage
\appendix

\section{Appendix / supplemental material}
\setcounter{theorem}{0}

\subsection{Effect of Batch Size}
\label{ap:batch-size}

Our analysis of quantization error in DP-SGD is independent of batch size. While larger batches reduce stochastic gradient variance, our argument hinges on the magnitude of the final noised gradient, which remains large regardless of batch size.

\begin{enumerate}
    \item In DP-SGD, the added noise’s scale is proportional to the per-example clipping constant, $C$, not the batch size.
    \item This noise dominates the averaged gradient signal, causing the raw gradients in subsequent steps to have significantly larger norms than in non-DP training.
    \item As shown in Proposition 1, quantization variance is proportional to the square of the gradient's norm ($\operatorname{Var}(q(x)) = \Theta(||x||_{\infty}^{2})$). Therefore, the larger gradients in DP-SGD lead to much higher quantization variance, which destabilizes training and degrades accuracy.
\end{enumerate}

To demonstrate this empirically, we ran the same training job with batch sizes ranging from 1024 to 8192 and measured the numerical range of the weight gradients, similar to that in figure 1c. Across the batch sizes, there is negligible difference in the gradient ranges, which confirms our hypothesis. 

\begin{table}[h!]
\centering
\caption{Weight gradient norm range across various batch sizes, showing the negligible impact of batch size on the final gradient magnitudes.}
\label{tab:gradient_norm_range}
\begin{tabular}{lcc}
\toprule
\textbf{Batch Size} & \textbf{Norm Range Mean} & \textbf{Norm Range Std} \\
\midrule
1024 & 0.159 & 0.137 \\
2048 & 0.161 & 0.127 \\
4096 & 0.158 & 0.116 \\
8192 & 0.156 & 0.119 \\
\bottomrule
\end{tabular}
\end{table}

\subsection{Evaluation Setup and Parameters}
\label{ap:dpquant-hyperparams}

\begin{table}[h]
    \centering
    \begin{tabular}{@{} l l p{0.6\linewidth} @{}}
        \toprule
        \textbf{Parameter} & \textbf{Default} & \textbf{Description} \\[3pt]
        $n$ & 60 & number of epochs to train \\
        $k$ & -- & layers to execute in low precision. \\
        $n_\text{sample}$ & 1 & test samples for loss measurement. \\
        $n_\text{interval}$ & 2 & epochs to train before the next measurement. \\
        $R$ & 2 & repetitions during measurement. \\  
        $\sigma_\text{measure}$ & 0.5 & Noise scale used during loss‐difference privatization. \\ 
        $\mathcal{C}_\text{measure}$ & 0.01 & Clipping norm used during loss‐difference privatization. \\
        \bottomrule
    \end{tabular}
  \caption{Configurable Hyperparameters of \codename}
\end{table}

\paragraph{Remark: Selecting \codename parameters in practice.} In our experiments, we have found repetitions = 2 and sampling frequency = 1 to be the most optimal. Adopting these recommended defaults, the user needs to pick:
\begin{enumerate}
    \item one of $k$ (number of layers to quantize) and the analysis frequency
    \item clipping norm used in loss sensitivity analysis
\end{enumerate}
The process of determining clipping norm for analysis is similar to that of finding the clipping threshold C for normal DP-SGD training. We want to pick a value such that the differences between policies are expressed.

\subsection{Evaluation under Extreme Privacy Budgets}
\label{ap:small-privacy-budget}
As shown in Figure~\ref{fig:privacy_consumption}, the privacy consumption of \codename's analysis accounts for a higher fraction of the total privacy near the beginning of training. We wish to evaluate \codename under more strict privacy budgets. 

In these cases, both the parameters of DP-SGD and \codename need to be updated, namely the noise scale $\sigma$ and measurement noise scale $\sigma_\text{measure}$ both need to be increased. Table~\ref{tab:acc-small-eps} below shows that \codename achieves optimal accuracy even when $\varepsilon=1$.

\begin{table}[h!]
\centering
\caption{Training accuracy of ResNet18 with GTSRB under the strict privacy budget ($\varepsilon=1$)}
\begin{tabular}{lcccc}
\toprule
& \multicolumn{2}{c}{\textbf{Baseline}} & \multicolumn{2}{c}{\textbf{\codename}} \\
\cmidrule(lr){2-3} \cmidrule(lr){4-5}
\textbf{Count} & \textbf{Accuracy (\%)} & \textbf{$\varepsilon$} & \textbf{Accuracy (\%)} & \textbf{$\varepsilon$} \\
\midrule
50\% & 44.14 $\pm$ 4.61 & 1.05 & 48.26 & 0.99 \\
75\% & 40.13 $\pm$ 3.65 & 1.05 & 43.14 & 0.99 \\
90\% & 35.20 $\pm$ 1.12 & 1.05 & 38.66 & 0.99 \\
\bottomrule
\end{tabular}
\label{tab:acc-small-eps}
\end{table}

\subsection{Training Hyperparameters}
\label{ap:train-hyperparams}

\subsubsection{Image Models}
While the learning rate might seem too high for regular SGD training, previous results~\cite{morsbach2024rrunderstandinghyperparametereffectsdpsgd,Ponomareva_2023} have shown that large learning rates are more beneficial for DP-SGD training.

\begin{table}[ht]
  \centering
  \small
  \caption{Experimental configurations (6 runs)}
  \label{tab:config}
  \begin{tabular}{lcccccc}
    \toprule
      & 1 & 2 & 3 & 4 & 5 & 6 \\
    \midrule
    Model                & ResNet18 & ResNet18 & ResNet18 & ResNet50 & DenseNet121 & DenseNet121 \\
    Dataset              & EMNIST & CIFAR10 & GTSRB & GTSRB & CIFAR10 & GTSRB \\
    $\sigma$             & $1$ & $1$ & $1$ & $1$ & $1$ & $1$ \\
    $\delta$             & $10^{-5}$ & $10^{-5}$ & $10^{-5}$ & $10^{-5}$ & $10^{-5}$ & $10^{-5}$ \\
    Clipping norm        & $1$ & $1$ & $1$ & $1$ & $1$ & $1$ \\
    Batch size           & $1024$ & $1024$ & $1024$ & $1024$ & $512$ & $512$ \\
    Physical batch size  & $128$ & $128$ & $128$ & $128$ & $128$ & $128$  \\
    Weights              & None  & ImageNet & ImageNet & ImageNet & ImageNet & ImageNet \\
    Optimizer            & SGD  & SGD & SGD & SGD & SGD & SGD \\
    Learning rate (lr)   & $0.5$ & $0.5$ & $0.5$ & $0.5$ & $0.5$ & $0.5$ \\
    Epochs               & $30$ & $60$ & $60$ & $60$ & $60$ & $60$ \\
    \bottomrule
  \end{tabular}
\end{table}

\subsubsection{Language Models}
We conducted a new NLP experiment using BERT for sequence classification on the Stanford Natural Language Inference (SNLI) corpus. In this task, the model classifies a pair of statements (e.g., “Children smiling and waving at camera” and “There are children present”) as “entailment,” “contradiction,” or “neutral.”

Due to the high number of parameters in BERT, we have followed the tutorial from Opacus and frozen 12 out of 13 BERT layers, and trained the last BERT layer and subsequent classification layers.

We have compared our method (\codename) with a random static baseline (similar to section 6.1). We use the same training parameters, trained for a single epoch, and used $\varepsilon=8$ as the total privacy budget.

In these experiments, \codename outperforms the baseline in accuracy. \codename consistently avoids quantizing the last few layers (including the trainable ones) without prior information about the importance and trainability of the layers.

\subsection{Evaluation with DP-Adam}
\label{ap:dp-adam}
We evaluate \codename using DP-Adam as the optimizer. All hyperparameters match the DP-SGD configuration in Table~\ref{tab:config}, with the optimizer changed to Adam (learning rate $0.01$, default $\beta_1=0.9$, $\beta_2=0.999$). 

Table~\ref{tab:dp-adam} shows that \codename consistently outperforms the static baseline under DP-Adam while consuming a strictly lower privacy budget, with the largest gains at 75\% and 90\% quantization.

\begin{table}[h!]
\centering
\caption{DP-Adam: \codename vs.\ static random baseline (ResNet18).}
\label{tab:dp-adam}
\begin{tabular}{llcccc}
\toprule
& & \multicolumn{2}{c}{\textbf{Baseline}} & \multicolumn{2}{c}{\textbf{\codename}} \\
\cmidrule(lr){3-4} \cmidrule(lr){5-6}
\textbf{Dataset} & \textbf{\shortstack{Percent\\Quantized}} & \textbf{Accuracy (\%)} & \textbf{$\varepsilon$} & \textbf{Accuracy (\%)} & \textbf{$\varepsilon$} \\
\midrule
\multirow{3}{*}{CIFAR-10}
 & 0.50 & $54.18 \pm 1.48$ & 5.05 & \textbf{55.03} & 4.97 \\
 & 0.75 & $49.86 \pm 1.57$ & 5.05 & \textbf{52.40} & 4.97 \\
 & 0.90 & $47.36 \pm 1.36$ & 5.05 & \textbf{51.18} & 4.97 \\
\midrule
\multirow{3}{*}{GTSRB}
 & 0.50 & $72.74 \pm 6.04$ & 6.05 & \textbf{75.87} & 5.91 \\
 & 0.75 & $58.14 \pm 7.64$ & 6.05 & \textbf{72.32} & 5.91 \\
 & 0.90 & $50.88 \pm 5.11$ & 6.05 & \textbf{66.45} & 5.91 \\
\bottomrule
\end{tabular}
\end{table}

Additionally, the BERT experiment on SNLI (Table~\ref{tbl:diff_eps}) uses DP-AdamW as the underlying optimizer.

\subsection{Accuracy Degradation for DP-SGD under Naive Quantization}
\label{ap:dp-degradation-other}
Prior works~\cite{ultralow,chmiel2024accurateneuraltraining4bit,micikevicius2022fp8formatsdeeplearning} have demonstrated minimal degradation during quantized fp4/8 training compared to the full precision counterpart. We tabulate their results below:

\begin{table}[h]
\centering
\caption{Ultra-Low and LUQ vs.\ baseline accuracy}
\label{tab:ulq-comparison}
\begin{tabular}{lccc}
\toprule
Model  & Baseline           & Ultra-Low~\cite{ultralow}           & LUQ~\cite{chmiel2024accurateneuraltraining4bit}                 \\
\midrule
ResNet-18        & 69.7\%          & 68.27\% (-1.43\%)        & 69.09\% (-0.61\%)         \\
ResNet-50        & 76.5\%          & 74.01\% (-2.49\%)        & 75.42\% (-1.08\%)         \\
MobileNet-V2     & 71.9\%          & 68.85\% (-3.05\%)        & 69.55\% (-2.35\%)         \\
ResNext50        & 77.6\%          & N/A                      & 76.02\% (-1.58\%)         \\
Transformer-base & 27.5 (BLEU)     & 25.4 (-2.10)             & 27.17 (-0.33)             \\
\bottomrule
\end{tabular}
\end{table}

As demonstrated in the Figure~\ref{fig:pareto}, the performance degradation of DP-SGD under quantization is much larger.

\begin{table}[ht]
\centering
\caption{Validation accuracy for DP-SGD training: baseline vs.\ LUQ-FP4 (all layers quantized)}
\label{tab:luq_fp4_summary}
\begin{tabular}{lllrrr}
\toprule
Model         & Dataset  & Pretraining            & Baseline & LUQ-FP4 & $\Delta$  \\
\midrule
ResNet-18     & EMNIST   & None    & 83.4\%   & 77.8\%   & -5.6\%   \\
ResNet-18     & CIFAR-10 & ImageNet           & 71.0\%   & 65.8\%   & -5.2\%   \\
ResNet-18     & GTSRB    & ImageNet           & 85.6\%   & 64.0\%   & -21.6\%  \\
ResNet-50     & GTSRB    & ImageNet           & 89.8\%   & 49.0\%   & -40.8\%  \\
DenseNet-121  & CIFAR-10 & ImageNet           & 67.0\%   & 62.9\%   & -4.1\%   \\
DenseNet-121  & GTSRB    & ImageNet           & 82.0\%   & 53.0\%   & -29.0\%  \\
\bottomrule
\end{tabular}
\end{table}

\subsection{Sensitivity of temperature $\beta$}
\label{ap:beta-sensitivity}
In our method, the temperature parameter $\beta$ provides a crucial mechanism to balance two complementary strategies:
\begin{itemize}
    \item Deterministic Selection: This approach prioritizes elements based on their loss sensitivity, selecting those that are most impactful.
    \item Randomized Sampling: This approach introduces stochasticity, ensuring diversity and exploration in the selection process.
\end{itemize}
A low $\beta$ value favors randomized sampling, while a high $\beta$ value makes the selection process more deterministic and reliant on loss sensitivity. Below, we tabulate the training accuracy for different value of $\beta$, and observe that better training outcomes can be obtained by favoring loss-based layer selection while retaining some stochasticity. Namely, it performs strictly better than selection purely based on random layer sampling.

\begin{table}[h!]
\centering
\caption{Model performance across various counts and temperature ($\beta$) values}
\label{tab:performance_data}
\sisetup{table-format=2.2} %
\begin{tabular}{l *{9}{S}}
\toprule
& \multicolumn{9}{c}{\textbf{Temperature ($\beta$)}} \\
\cmidrule(lr){2-10}
\textbf{Count} & {0.1} & {0.22} & {0.47} & {1.03} & {2.24} & {4.86} & {10.57} & {22.99} & {50.0} \\
\midrule
10 & 66.49 & 67.58 & 67.53 & 67.01 & 70.25 & 70.37 & \textbf{71.59} & 70.96 & \textbf{71.67}\\
15 & 58.47 & 58.47 & 60.03 & 59.07 & 60.86 & \textbf{65.00} & 60.54 & \textbf{65.04} & 63.75 \\
18 & 51.60 & 54.08 & 55.73 & 53.73 & 53.86 & 53.49 & \textbf{60.90} & 55.45 & 56.05 \\
\bottomrule
\end{tabular}
\end{table}

\subsection{EMA Ablation}
\label{ap:ema-ablation}
The exponential moving average (EMA) in step~\blackcircled{4} of Algorithm~\ref{alg:compute_regret_concise} smooths the per-layer importance scores across measurement iterations so that a single noisy batch does not drastically change the layer ranking. Table~\ref{tab:ema-ablation} shows the effect of removing EMA on ResNet18 trained on CIFAR-10 with DP-SGD. Across all quantization levels, EMA consistently improves accuracy, confirming that smoothing the noisy sensitivity estimates is beneficial.

\begin{table}[h!]
\centering
\caption{Effect of EMA on \codename (ResNet18, CIFAR-10, DP-SGD).}
\label{tab:ema-ablation}
\begin{tabular}{lcc}
\toprule
\textbf{Percent Quantized} & \textbf{With EMA} & \textbf{Without EMA} \\
\midrule
0.50 & 56.06 & 54.01 \\
0.75 & 51.59 & 50.07 \\
0.90 & 49.97 & 47.22 \\
\bottomrule
\end{tabular}
\end{table}

\subsection{Evaluation on other Quantizers}
\label{ap:other-quantizers}
To assess the versatility of our method, DPQuant, we evaluated its performance with different numerical precisions and quantization schemes. We conducted two primary experiments: one using 8-bit floating-point (\texttt{fp8e5m2}) for training to test a different bitwidth, and another using a uniform 4-bit quantizer to test a different quantization strategy.

\subsubsection{FP8 Quantization}
Our experiments with FP8 training show that quantized DP-SGD does not suffer a significant performance degradation. This minimal performance gap suggests that in higher-precision settings like FP8, the benefits of more complex techniques like layer subset selection may be less critical. We show the results in table~\ref{tab:fp8_performance}.
\begin{table}[h!]
\centering
\caption{Performance comparison with FP8 training.}
\label{tab:fp8_performance}
\begin{tabular}{lcccc}
\toprule
\textbf{Count} & \textbf{Base Acc(\%)} & \textbf{Base $\epsilon$} & \textbf{Our Acc(\%)} & \textbf{Our $\epsilon$} \\
\midrule
50\% & $67.56 \pm 0.47$ & 4.05 & 67.12 & 3.93 \\
75\% & $67.76 \pm 0.67$ & 4.05 & 67.65 & 3.93 \\
90\% & $67.38 \pm 0.59$ & 4.05 & 68.01 & 3.93 \\
\bottomrule
\end{tabular}
\end{table}

\subsubsection{Uniform 4-bit Quantization}

Next, we evaluated a more aggressive quantization scheme using a uniform FP4 quantizer. In this setup, the value range is discretized into $2^4=16$ levels via stochastic rounding. The results reveal a more substantial drop in accuracy for our method compared to the baseline. This outcome is consistent with our observations of the LUQ-FP4 quantizer discussed in Section 6.2, highlighting the inherent challenges of applying DP-SGD with very low-bitwidth uniform quantization. We show the results in table~\ref{tab:fp4_performance}.

\begin{table}[h!]
\centering
\caption{Performance comparison with uniform FP4 quantization.}
\label{tab:fp4_performance}
\begin{tabular}{lcccc}
\toprule
\textbf{Count} & \textbf{Base Acc(\%)} & \textbf{Base $\epsilon$} & \textbf{Our Acc(\%)} & \textbf{Our $\epsilon$} \\
\midrule
50\% & $63.56 \pm 0.89$ & 4.53 & 62.15 & 4.44 \\
75\% & $57.85 \pm 0.90$ & 4.53 & 59.09 & 4.44 \\
90\% & $55.82 \pm 0.80$ & 4.53 & 56.27 & 4.44 \\
\bottomrule
\end{tabular}
\end{table}

\subsection{Proof of Proposition 1}
\label{ap:proof-prop-1}
\begin{theorem}
Let \(q:\mathbb{R}^n\to\mathbb{R}^n\) be an unbiased (\(\mathbb{E}[q(\mathbf{x})]=\mathbf{x}\)) and scale‐invariant ($q(\lambda\mathbf{x})=\lambda\,q(\mathbf{x})$) quantizer whose outputs lie on a fixed finite grid.  If \(\mathbf{x}\) is drawn from an absolutely continuous distribution, then
\[
\operatorname{Var}\bigl(q(\mathbf{x})\bigr)\;=\;\Theta\bigl(||\mathbf{x}||_\infty^2\bigr).
\]

\begin{proof}
We begin by first showing the upper-bound: $\operatorname{Var}\left( q(\mathbf{x}) \right)= \mathcal{O}\left( ||\mathbf{x}||_\infty^2 \right)$. We define \(M=||\mathbf{x}||_\infty\) and \(\mathbf{v}=\mathbf{x}/M\), so \(||\mathbf{v}||_\infty=1\).  By scale‐invariance of \(q\),
\[
\operatorname{Var}\bigl(q(\mathbf{x})\bigr)
=\operatorname{Var}\bigl(q(M\mathbf{v})\bigr)
=\operatorname{Var}\bigl(M\,q(\mathbf{v})\bigr)
=M^2\,\operatorname{Var}\bigl(q(\mathbf{v})\bigr).
\]
Since \(q(\mathbf{v}) \in [-1,1]^n\), there exists a finite $C$ such that
\(\operatorname{Var}(q(\mathbf{v}))\le C\), giving
\[
\operatorname{Var}\bigl(q(\mathbf{x})\bigr)=M^2\operatorname{Var}\left(q(\mathbf{v})\right) \le C\,M^2 = C\,||\mathbf{x}||_\infty^2.
\]
Next, we show the lower-bound: $\operatorname{Var}\left( q(\mathbf{x}) \right)=\Omega\left( ||\mathbf{x}||_\infty^2 \right)$. On the compact set \(\{\mathbf{v}:||\mathbf{v}||_\infty=1\}\), the continuous function \(v\mapsto \operatorname{Var}(q(\mathbf{v}))\) attains a minimum \(m\ge0\).  Because the finite quantizer grid has measure-zero, absolute continuity of \(\mathbf{x}\) ensures the probability of $\mathbf{v}$ landing on the grid is $0$, so \(\operatorname{Var}(q(\mathbf{v}))>0\) and hence \(m>0\). Therefore
\[
m\,M^2 \;\le\;\operatorname{Var}\bigl(q(\mathbf{x})\bigr)\;\le\;C\,M^2,
\]
From this we conclude \(\operatorname{Var}(q(\mathbf{x}))=\Theta(||\mathbf{x}||_\infty^2)\), as desired.
\end{proof}
\end{theorem}

\subsection{Justification of Privacy Guarantees (Theorem 2)}
\label{ap:proof-prop-2}

\begin{theorem}
    Algorithm \ref{alg:compute_regret_concise} is a Sampled Gaussian Mechanism (SGM) with sample rate $q=|B|/|D|$ and noise scale $\sigma=\sigma_\text{measure}$. 
    \label{prop:sgm-appendix}
    \begin{proof}
        We first characterize Algorithm~\ref{alg:compute_regret_concise} as an analysis on the user's private dataset. The function accepts a subsampled batch of size $|B|$ from a dataset with $|D|$ samples. 

        Using this batch of user data, as well as some non-private sources of data such as the model weights, we compute the loss differences which is vectorized in $\mathbf{R}\in \mathbb{R}^p$, where $p$ is the number of available quantization policies.  

        In step~\blackcircled{3} of Algorithm~\ref{alg:compute_regret_concise}, we clip the vector $\mathbf{R}$ to norm $\mathcal{C}$, to which independent Gaussian noise proportional to $\sigma^2\mathcal{C}^2$ is added to obtain $\hat{\mathbf{R}}$. This is equivalent to adding noise proportional to $\sigma^2$ when the sensitivity of $\hat{\mathbf{R}}$ is $1$ through a scaling argument.

        Algorithm~\ref{alg:compute_regret_concise} ceases to access private data in $B$ after the computation of $\hat{\mathbf{R}}$, which results in step~\blackcircled{4} (updating the policy EMA) being post-processing~\cite{dwork2014algorithmic} which does not impact the privacy consumption. 

        Furthermore, the privacy accounting step makes use of Opacus'~\cite{yousefpour2022opacususerfriendlydifferentialprivacy} privacy accountant, which assumes\footnote{This assumption is stated in \texttt{https://github.com/pytorch/opacus/blob/main/opacus/accountants/analysis/rdp.py}} the noise scale $\sigma$ is proportional to the clipping constant (i.e. equivalent to adding a noise proportional to $\sigma^2\mathcal{C}^2$. 
    \end{proof}
\end{theorem}

\subsection{Low Precision Simulation Setup}
As FP4 hardware support is forthcoming, we employ the following simulation setup to emulate the effect of training under FP4. Notably, we quantize both inputs to the conv2d forward, wgrad, and dgrad operators as well as its output. 

\begin{figure}[h!]
    \centering
    \includegraphics[width=0.65\linewidth]{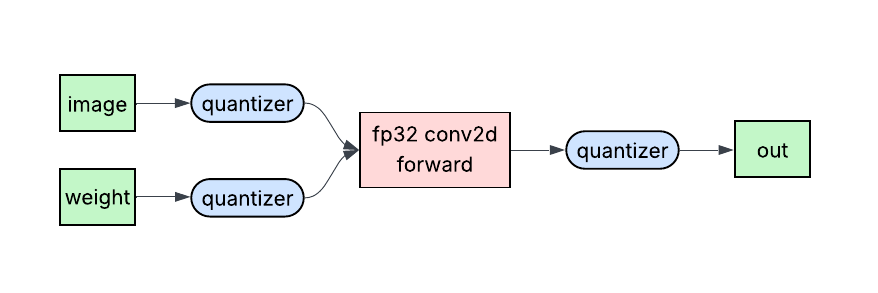}
    \caption{Quantization simulation setup}
    \label{fig:simulation}
\end{figure}
\subsection{Theoretical Speedup Calculation}
\label{ap:speedup}
Due to the unavailability of accelerators and reliable software support for FP4, we instead rely on a performance model to estimate the theoretical throughputs of FP4 computation. 

We first decompose the DP-SGD training computation into the following operations, listed in table~\ref{tab:dpsgd-computation-stage}. 

\begin{figure}[h!]
    \centering
    \includegraphics[width=\linewidth]{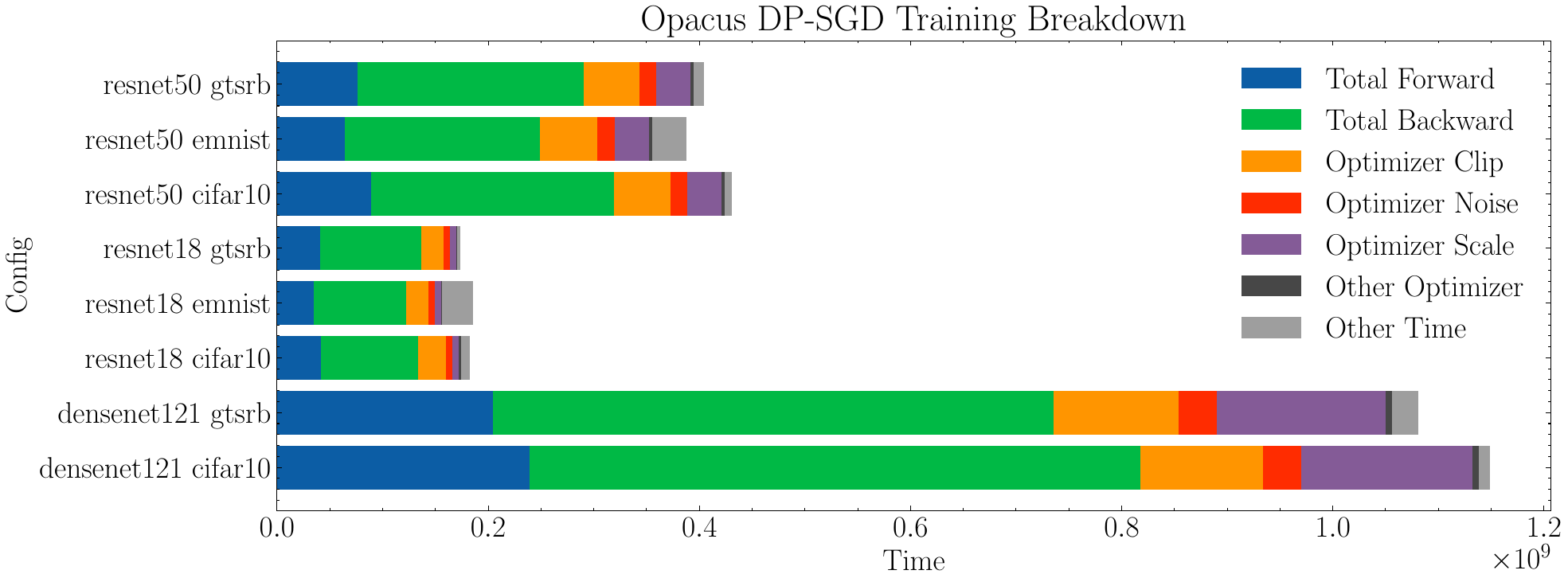}
    \caption{Runtime decomposition of DP-SGD training}
    \label{fig:decomposition}
\end{figure}

We performed profiling on the models (on their respective datasets) stated in the paper, and we plot the runtime decomposition in Figure~\ref{fig:decomposition}. Using this data, we can compute the amount of ``overhead'' (i.e. the time spent on operators which will not benefit from lower precision) for each model/dataset. This is tabulated in Table~\ref{tab:overhead_analysis}.

\begin{table}[h!]
\centering
\caption{Decomposition of DP-SGD training}
\label{tab:dpsgd-computation-stage}
\begin{tabular}{@{}l p{0.6\textwidth} c@{}}
\toprule
\textbf{Computation Stage} & \textbf{Description} & \textbf{\shortstack{Benefits\\from FP4}} \\
\midrule
\emph{Total Forward} 
  & Time spent on the forward pass through the model, where input data is processed layer by layer to produce the output. 
  & \checkmark \\
\addlinespace
\emph{Total Backward} 
  & Time for backpropagation, where gradients are calculated for model parameter updates. 
  & \checkmark \\
\addlinespace
\emph{Optimizer Clip} 
  & Time for clipping gradients to a predefined threshold to ensure stability and prevent large updates during training. 
  & \checkmark \\
\addlinespace
\emph{Optimizer Noise} 
  & Time for adding random noise to the gradients to ensure differential privacy by masking individual data point contributions. 
  &  \\
\addlinespace
\emph{Optimizer Scale} 
  & Time for scaling the gradients after clipping to adjust the magnitude of the updates. 
  & \checkmark \\
\addlinespace
\emph{Other Optimizer} 
  & Time spent on other optimizer-related operations, such as learning rate management. 
  &  \\
\addlinespace
\emph{Other Time} 
  & Time for all other operations during the training iteration, including data loading, synchronization, and auxiliary tasks. 
  &  \\
\bottomrule
\end{tabular}
\vspace{5pt}
\end{table}

\newcommand{\scinum}[1]{\num[scientific-notation = true, round-mode=figures, round-precision=3]{#1}}

\begin{table}[h!]
\centering
\caption{Breakdown of total time, good ops, bad ops, and overhead percentage for different model configurations.}
\label{tab:overhead_analysis}
\begin{tabular}{@{}lrrrr@{}}
\toprule
\textbf{Config}             & \textbf{Total Time}  & \textbf{Ops with Speedup}   & \textbf{Overhead Ops}    & \textbf{Overhead \%} \\ \midrule
DenseNet121 CIFAR10 & $\scinum{1148963858.1960497}$ & $\scinum{1096631480.596805}$ & $\scinum{52332377.599244595}$ & 4.55 \\
DenseNet121 GTSRB & $\scinum{1081264406.9638088}$ & $\scinum{1013885545.0322825}$ & $\scinum{67378861.9315263}$ & 6.23 \\
ResNet18 CIFAR10 & $\scinum{182280989.2413641}$ & $\scinum{165516083.9176193}$ & $\scinum{16764905.323744804}$ & 9.20 \\
ResNet18 EMNIST & $\scinum{185872767.91796964}$ & $\scinum{149055829.99314263}$ & $\scinum{36816937.92482701}$ & 19.81 \\
ResNet18 GTSRB & $\scinum{173634863.2583754}$ & $\scinum{163229706.0868052}$ & $\scinum{10405157.171570212}$ & 5.99 \\
ResNet50 CIFAR10 & $\scinum{430607249.5522515}$ & $\scinum{405131911.34800273}$ & $\scinum{25475338.204248786}$ & 5.92 \\
ResNet50 EMNIST & $\scinum{387656456.4085218}$ & $\scinum{336396564.6420171}$ & $\scinum{51259891.766504645}$ & 13.22 \\
ResNet50 GTSRB & $\scinum{404541737.62423193}$ & $\scinum{375801335.094198}$ & $\scinum{28740402.530033946}$ & 7.10 \\

\bottomrule
\end{tabular}
\end{table}

\newpage

\subsection{Opacus Privacy Accounting}
\label{ap:opacus-accounting}
Opacus maintains a tuple of the form (sample rate, noise scale, number of steps), which is incrementally updated during training. At any point, we can query the current privacy cost in terms of $(\varepsilon, \delta)$ by specifying a target $\delta$ and using either the \texttt{rdp} or \texttt{prv} accountant. This mechanism enables flexible and precise tracking of privacy usage, allowing us to assess how much additional privacy is consumed by our analysis relative to standard training.
\subsection{Sampling of Quantized Layers}
\label{ap:select-targets}
We outline the algorithm \codename uses to select layers to compute under quantization in Algorithm~\ref{alg:select-targets}. 
\begin{algorithm}[h!]
\caption{\textsc{SelectTargets}}
\begin{algorithmic}[1]
\State \textbf{Input:} $L$ (EMA scores), $P$ (set of policies), $s$ (temperature), $m$ (number to sample), \texttt{layers} (set of layers to quantize under policy $p$)

\State $v \gets [L[p] \text{ for } p \in P]$
\State $v \gets (v - \min(v)) / (\max(v) - \min(v))$ \Comment{Normalize}
\State $\pi \gets \texttt{softmax}(-s \cdot v)$
\State $Q \gets \text{Multinomial}(\pi,\, m,\, \text{without replacement})$ \Comment{Sample $m$ policies}

\State $S \gets \emptyset$
\For{each $p \in Q$}
    \State $S \gets S \cup \texttt{layers}[p]$
\EndFor

\State \Return $S$
\end{algorithmic}
\label{alg:select-targets}
\end{algorithm}

\subsection{Remark: Applicability to Non-DP Training}
\label{ap:non-dp}

The mechanisms in Sections~\ref{sec:part1} and~\ref{sec:part2} are general and could in principle apply to non-DP settings. However, prior work such as LUQ-FP4~\citep{chmiel2024accurateneuraltraining4bit} shows that FP4 non-DP training suffers negligible degradation, so \codename's improvements would be marginal there. Our main novelty and motivation are DP-specific: we identify and analyze a setting where non-DP FP4 training shows negligible degradation, but DP + FP4 suffers substantial accuracy loss (Figure~\ref{fig:nondp_accuracy}), attributing this to the interaction between $\ell_2$-calibrated DP noise and $\ell_\infty$-driven quantization. Furthermore, our loss-sensitivity measures and selection rules are designed to be compatible with DP accounting and composition, relying only on quantities that are already DP-released or incorporable into the privacy accountant, which is not a concern in standard non-DP quantization work.

\subsection{Remark: Vulnerability to Floating Point Attacks}
Differential privacy implementations must carefully consider the vulnerabilities highlighted by \cite{10.1145/2382196.2382264}. Mironov identified that the floating-point implementation of noise sampling for mechanisms such as Laplacian or Gaussian introduces a “porous” distribution that lacks translation invariance. This issue is prevalent in both fp64 and fp32 arithmetic.

To ensure robustness against this vulnerability, our method has been meticulously designed. The critical step of noise addition in our framework occurs under standard conditions, prior to the application of our novel quantization technique. The process is as follows:

\begin{enumerate}
    \item Gradients are maintained in full fp32 precision.
    \item Noise is sampled and added to these fp32 gradients, also in fp32 precision.
    Only after the noisy gradient is computed is it quantized for use in the forward/backward pass of select layers.
    \item Thus, the noise injection process maintains a vulnerability profile identical to that of standard DP-SGD implemented in fp32. The use of lower-precision representations for computation does not alter or exacerbate the known properties of the initial noise addition.

\end{enumerate}

Additionally, our method is fully compatible with established defenses against this vulnerability. The 'snapping mechanism' proposed by Mironov, a post-processing step applied directly to the noisy output, would be applied to the full-precision fp32 gradients immediately after noise addition and before quantization in our pipeline.